\documentclass{article}
\usepackage{ijcai25}

\usepackage{times}
\usepackage{soul}
\usepackage{url}
\usepackage[hidelinks]{hyperref}
\usepackage[utf8]{inputenc}
\usepackage[small]{caption}
\usepackage{graphicx}
\usepackage{caption} 
\usepackage{amsmath}
\usepackage{amsthm}
\usepackage{booktabs}
\usepackage{algorithm}
\usepackage{algorithmic}
\usepackage{bbding}
\usepackage{booktabs}
\usepackage{hyperref}
\usepackage{amssymb}

\usepackage[dvipsnames]{xcolor}

\onecolumn
\renewcommand\twocolumn[1][]{#1}%

\title{CLGRPO: Reasoning Ability Enhancement for Small VLMs}

\author{
Fanyi Wang$^*$\textsuperscript{\rm 1},
Binzhi Dong\textsuperscript{\rm 1},
Haotian Hu\textsuperscript{\rm 2},
Jinjin Xu\textsuperscript{\rm 3},
Zhiwang Zhang\textsuperscript{\rm 4}
\affiliations
Honor AI Center\textsuperscript{\rm 1} \and Zhejiang University\textsuperscript{\rm 2} \and Bytedance\textsuperscript{\rm 3}\and Ningbo Tech University\textsuperscript{\rm 4}\\
\emails
{\{wangfanyi,dongbinzhi\}}@honor.com \and hht1996ok@zju.edu.cn \and \\  jin.xu@mail.ecust.edu.cn \and zhiwang.zhang@nit.zju.edu.cn
}

\begin{document}
\maketitle

\begin{abstract}
Small Vision Language Models (SVLMs) generally refer to models with parameter sizes less than or equal to 2B. Their low cost and power consumption characteristics confer high commercial value. However, their reasoning abilities are limited by the number of parameters. To address this issue, this paper proposes a post-training optimization paradigm called the Incremental Training Strategy to enhance the reasoning ability of SVLMs. Firstly, we constructed a Self-Supervised Chain-of-Thought (COT) Data Construction System, which leverages multiple LVLMs with 7B parameters or more to transform original data into COT data in a self-supervised manner. Our proposed Incremental Training Strategy consists of four stages. Stage 1 injects domain knowledge by performing Supervised Fine-Tuning (SFT) to the pretrained model on the COT data. Stage 2 aligns the COT data format by conducting a small amount of Group Relative Policy Optimization (GRPO) training constrained only by format rewards on the COT data. Stage 3 enhances reasoning ability by applying GRPO training on the COT data with constraints on both format and accuracy rewards. The resulting model shows significant improvement compared to the baseline. Stage 4 addresses the limited capacity of the SVLMs and the weak ability to capture complex patterns by proposing ClipLow GRPO (CLGRPO) to constrain the capture space of the training process. We conducted extensive comparative and ablation experiments on the abstract semantic recognition dataset EMOSet-118K. Experimental results demonstrate that our method significantly improves the reasoning ability of 1B SVLM. Compared to the baseline model fine-tuned on the original data, accuracy increased by 2.77 and recall by 0.69, achieving performance comparable to that of 8B models.
\end{abstract}

\section{Introduction}

With the success of the Qwen series \cite{Qwen-VL}\cite{Qwen2-VL}\cite{Qwen2.5-VL}\cite{yang2025qwen3}, InternVL series \cite{chen2024internvl}\cite{chen2024internvl1.5}\cite{chen2024expanding}\cite{zhu2025internvl3exploringadvancedtraining}, and DeepSeek series \cite{deepseekai2024deepseekv3technicalreport}\cite{deepseekai2025deepseekr1incentivizingreasoningcapability}\cite{lu2024deepseekvl}\cite{wu2024deepseekvl2mixtureofexpertsvisionlanguagemodels} Vision Language Models in both academia and industry, an increasing number of researchers have joined the wave of VLM research\cite{xu2023u}\cite{xu2025reproducibility}\cite{wang2025overcoming}. Model lightweighting and miniaturization~\cite{fastvlm2025}~\cite{zhang2025flashvl} to adapt to edge-side private and personalized deployment have become one of the research hotspots. For terminal manufacturers, small Vision Language Models (SVLMs) with parameter sizes less than or equal to 2B possess extremely high commercial value due to their relatively low cost, low power consumption, and customizable attributes. However, SVLMs have limited model capacity and relatively weak reasoning capabilities, and currently, there are few targeted methods proposed to enhance the reasoning ability of SVLMs.
To address this situation, we propose a post-training optimization paradigm called the Incremental Training Strategy to improve the reasoning ability of SVLMs. We select the abstract semantic understanding task as the entry point and conduct experiments on the EmoSet-118K dataset~\cite{EmoSet}, choosing InternVL-1B~\cite{chen2024internvl} as the baseline model. We construct a Self-Supervised Chain-of-Thought (COT) Data Construction system that uses multiple LVLMs with 7B parameters or more to convert the EmoSet-118K dataset into COT-formatted data, which is then self-supervisedly verified. Our proposed Incremental Training Strategy consists of four main training stages.
Stage 1: Perform SFT to the pretrained model on the COT data to inject prior knowledge. The model obtained in this step performs worse on metrics compared to the baseline model fine-tuned on the original data.
Stage 2: Based on the model from Stage 1, conduct GRPO fine-tuning on the COT data, constrained only by the format reward to achieve format alignment.
Stage 3: Use GRPO to simultaneously constrain both format reward and accuracy reward, aiming to enhance reasoning ability. The model obtained in this step shows significant improvement over the baseline, indicating that after stepwise knowledge injection and reasoning enhancement, SVLM preliminarily possesses the ability to further improve through extended reasoning.
Stage 4: To address the small capacity and weak complex pattern capture ability of SVLMs, we further propose the ClipLow GRPO method to constrain the SVLMs’ capture space. Experimental results demonstrate that our proposed CLGRPO further benefits the reasoning ability of SVLMs, with significant metric improvements over Stage 3. Our contributions are as follows.
\begin{itemize}
\item We designed a Self-Supervised COT Data Construction System that automatically processes the EmoSet-118K dataset into high-quality COT-formatted data. 
\item We propose a four-stage post-training optimization paradigm, the Incremental Training Strategy, to enhance the reasoning ability of SVLMs.
\item To address the limited capacity and weak complex pattern capture ability of SVLMs, we introduce the ClipLow GRPO method, which constrains the SVLM’s capture space during training. On the EmoSet-118K benchmark, this approach improves domain-specific capabilities of SVLM, achieving performance comparable to that of LVLMs with eight times the number of parameters after SFT.
\end{itemize}
\section{Related Work}

With the continuous enhancement of Vision-Language Model (VLM) capabilities and knowledge density, these models have been increasingly applied across a wide range of domains. Among the currently open-source VLM series, the leading ones in terms of performance are primarily the InternVL series \cite{chen2024internvl}\cite{chen2024internvl1.5}\cite{chen2024expanding}\cite{zhu2025internvl3exploringadvancedtraining}and Qwen series \cite{Qwen-VL}\cite{Qwen2-VL}\cite{Qwen2.5-VL}\cite{yang2025qwen3}. The core contribution of InternVL1.0~\cite{chen2024internvl} was to scale the visual encoder’s parameter size to be comparable with that of the text encoder. InternVL1.1 focused on improving Chinese language understanding and OCR capabilities. InternVL1.2 integrated multiple models to strengthen the visual encoder and replaced the text encoder with the larger Nous-Hermes-2-Yi-34B. InternVL1.5~\cite{chen2024internvl1.5} employed a more powerful visual encoder and upgraded the text encoder to InternLM2-20B. InternVL2 replaced the heavy 6B visual encoder with a lightweight 300M ViT, covering a parameter range from 2B to 108B, where the 2B model enables potential deployment on edge devices. InternVL2.5 \cite{chen2024expanding} introduced JPEG compression and loss function reweighting techniques. InternVL3~\cite{zhu2025internvl3exploringadvancedtraining} further incorporated variable visual encoding, native multimodal pretraining, and hybrid preference optimization modules, achieving superior long-context understanding capabilities.

The success of the Qwen series relies on several key technologies, including the autoregressive generation mechanism, multi-head self-attention, feed-forward neural networks (FFN) with residual connections, as well as the implementation of positional encoding and input embeddings. These techniques not only enhance the model’s generative ability but also optimize efficiency and scalability. The Qwen1.5 series includes models of various scales, with the largest reaching 110B parameters. Some models adopt Grouped Query Attention (GQA) to improve inference speed and reduce memory consumption. The Qwen2 series\cite{Qwen2-VL} represents a significant upgrade, supporting much longer context lengths up to 128K tokens. This series comprises multiple pretrained and instruction-tuned models, such as Qwen2-0.5B, Qwen2-1.5B, Qwen2-7B, Qwen2-57B-A14B, and Qwen2-72B. All model sizes utilize GQA technology to further enhance inference efficiency. Building upon Qwen2, the Qwen2.5 series\cite{Qwen2.5-VL} extends multimodal capabilities and supports even larger context windows, releasing high-performance models like Qwen2.5-72B-Instruct. The latest Qwen3 series\cite{yang2025qwen3} further advances inference capabilities and multimodal fusion, enabling support for larger-scale context inputs and more complex task handling.


\subsection{Efficient Vision Language Models}

As VLM technology matures, there is an increasing demand for lightweight and personalized edge deployment, which has sparked a surge of research on efficient VLMs. Xiangxiang Chu et al. proposed MobileVLM~\cite{chu2023mobilevlm}, a multimodal vision-language model tailored for edge scenarios. Their lightweight downsampling projector, LDPv2, compresses the number of image tokens to 1/9 of traditional approaches, effectively addressing the issue of visual feature dimensionality explosion. FastVLM~\cite{fastvlm2025} introduced a novel hybrid visual encoder, FastViTHD, which employs a hybrid hierarchical design and multi-scale feature fusion techniques. This approach significantly reduces the encoding time for high-resolution images while maintaining the accuracy of the visual encoder. BlueLM-V-3B~\cite{2023bluelm} proposed a relaxed aspect ratio matching method, enabling more efficient inference on mobile devices. Flash-VL2B~\cite{zhang2025flashvl} utilizes implicit semantic concatenation to solve the semantic discontinuity problem in image patch processing, greatly enhancing the model’s performance in tasks such as document understanding and OCR.


\subsection{Reinforcement Learning}

Reinforcement learning~\cite{schulman2017proximal}~\cite{rafailov2023direct}~\cite{su2025reveal}, as a means to further enhance the capabilities of VLMs, has also attracted extensive research attention. DeepSeekMath introduced GRPO~\cite{deepseek-math}, which directly uses the average reward of multiple sampled outputs as the baseline without requiring an additional value function approximation, significantly reducing the training resources needed. Dr.GRPO~\cite{liu2025understanding} optimizes GRPO by addressing length bias and difficulty bias introduced during the optimization process, it removes the normalization factor for response length and instead employs Monte Carlo average return to estimate the advantage value, enabling the model to balance optimization across tasks of varying difficulty. DAPO~\cite{yu2025dapo} significantly improves training efficiency, stability, and output quality of large models on complex tasks by decoupling the upper and lower clipping ranges, dynamically sampling effective gradient data, introducing token-level policy optimization, and applying a soft penalty mechanism.


\begin{figure}[ht]
\begin{center}
    \captionsetup{type=figure}
    \includegraphics[width=1.0\linewidth]{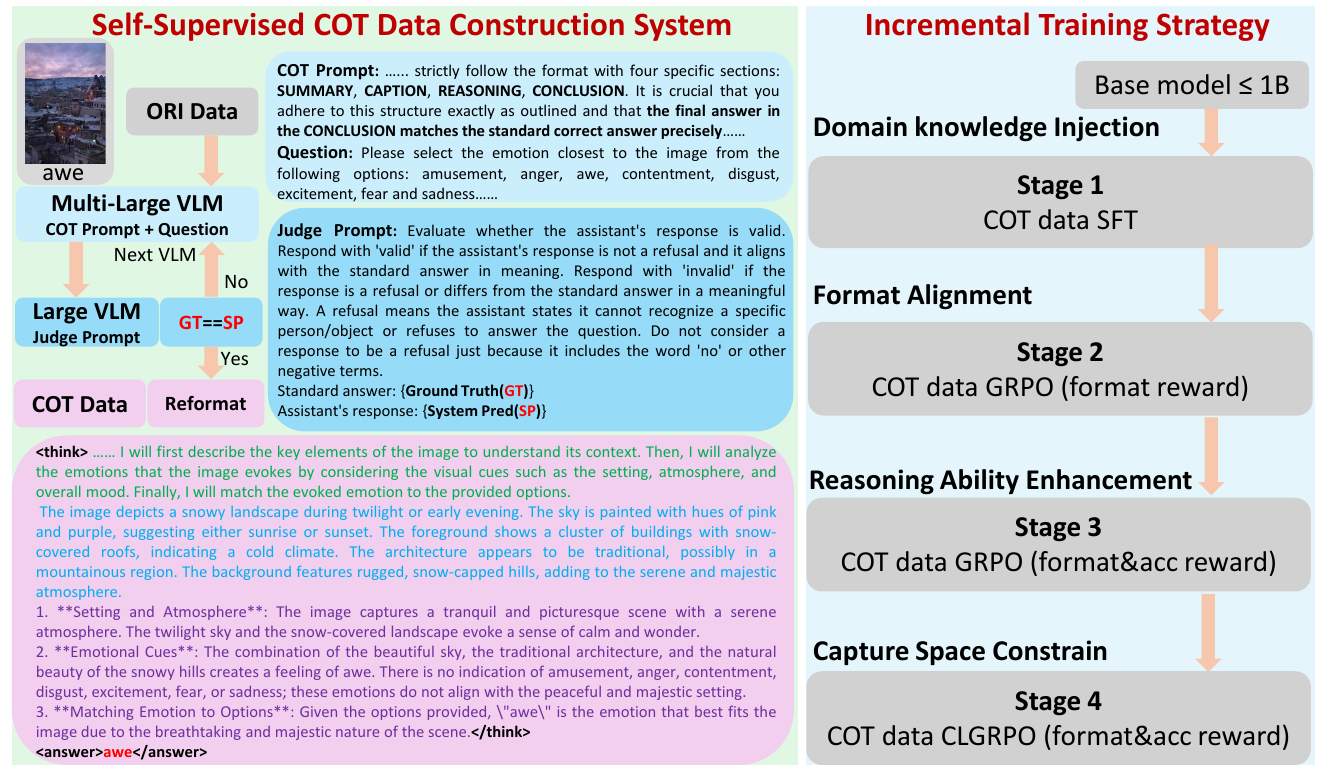}
    \captionof{figure}{Overview of \textbf{Self-Supervised COT Data Construction System} and our \textbf{Incremental Training Strategy}. \textcolor{Green}{Green is SUMMARY part}, \textcolor{Cyan}{Blue is CAPTION part}, \textcolor{Purple}{Purple is RESASONING part}, \textcolor{Red}{Red is CONCLUSION part}.}
    \label{fig:Overview}
\end{center}%
\end{figure}

\section{Methods}

Extensive research has shown that Chain-of-Thought (COT) reasoning can effectively enhance the reasoning capabilities of Large Vision Language Models (LVLMs). However, there is relatively limited research on COT for small Vision Language Models (SVLMs) with parameter sizes less than or equal to 2B~\cite{shao2024visual}~\cite{zhao2025cot}. Many perspectives suggest that the COT ability of SVLMs is constrained by their number of parameters and knowledge density, making it difficult to fully activate.

To address this issue, based on extensive experience, this paper explores a multistage training strategy that fundamentally and effectively improves the COT reasoning ability of SVLMs on domain-specific tasks.

Our experiments are conducted on the image abstract semantic recognition task, using the EmoSet-118K~\cite{EmoSet} as benchmark. We first design a Self-Supervised COT Data Construction System based on multiple large VLMs to automatically process EmoSet-118K into COT-formatted data. The specific processing procedure is illustrated on the left side of Figure.~\ref{fig:Overview} and will be introduced in Section 3.1. Subsequently, we perform a four-stage Incremental Training Strategy based on the COT data. The detailed steps are shown on the right side of Figure.~\ref{fig:Overview} and will be introduced in Section 3.2.

\subsection{Self-Supervised COT Data Construction System}
\label{sec:3.1}
\begin{figure}[t]
\begin{center}
    \captionsetup{type=figure}
    \includegraphics[width=1.0\linewidth]{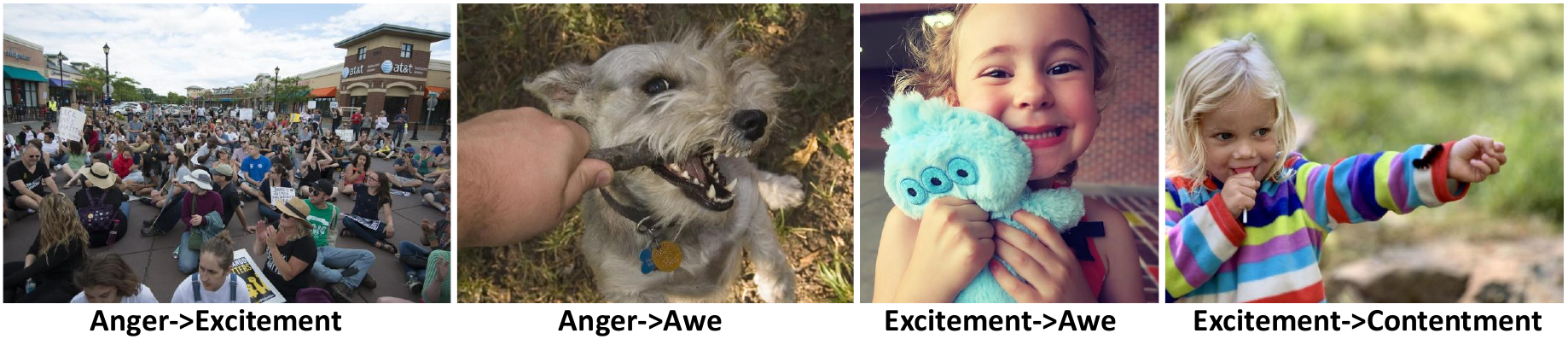}
    \captionof{figure}{Failure cases of Self-Supervised COT Data Construction System. The label bellow image above is "Ground Truth-$>$System Pred". The emotions expressed in these images are to some extent really easy to confuse.}
    \label{fig:Failure cases}
\end{center}%
\end{figure}

The benchmark we selected, EmoSet-118K~\cite{EmoSet}, presents certain challenges in the field of image abstract semantic recognition. The original dataset consists of pairs of image-emotion annotations, with emotions finely categorized into eight classes: sadness, amusement, awe, disgust, anger, excitement, fear, and contentment. We constructed a Self-Supervised Chain-of-Thought (COT) Data Construction System based on LVLMs. As illustrated on the left side of Figure.~\ref{fig:Overview}, the entire self-supervised process is divided into two stages.
In the first stage, LVLMs are guided jointly by a COT Prompt and a Question to generate data with reasoning steps. The inputs to this process include images and emotion labels from the original dataset. The COT Prompt instructs the LVLM to generate responses following the "SUMMARY, CAPTION, REASONING, CONCLUSION" sequences, and the generated conclusion must be consistent with the emotion label.
In the second stage, a Judge Prompt is used to have the large VLM verify whether the generated response’s CONCLUSION(System Pred) matches the Ground Truth. If inconsistency is detected, the process returns to the first stage, where a larger VLM is employed for regeneration. The pool of LVLMs used in the first stage includes InternVL2.5-7B/32B/72B-Instruct, called iteratively from smaller to larger. Considering processing efficiency and the relatively lower difficulty of the second stage, InternVL2.5-7B-Instruct is chosen for verification.
After the second stage verification, the generated results are reformatted into the following structure:"$<$think$>$\textcolor{Green}{SUMMARY},\textcolor{Cyan}{CAPTION},\textcolor{Purple}{REASONING}$<$/think$><$answer$>$\textcolor{Red}{CONCLUSION}$<$/answer$>$".
In practice, most data are annotated successfully after one round of the first stage. After three rounds of iterative self-supervision, 78 data samples were filtered out by the Self-Supervised COT Data Construction System. Upon inspection of the filtered data, as shown in Figure.~\ref{fig:Failure cases}, we think it may well be caused by the fact that those data are easy to confuse.

\begin{figure}[ht]
\begin{center}
    \captionsetup{type=figure}
    \includegraphics[width=1.0\linewidth]{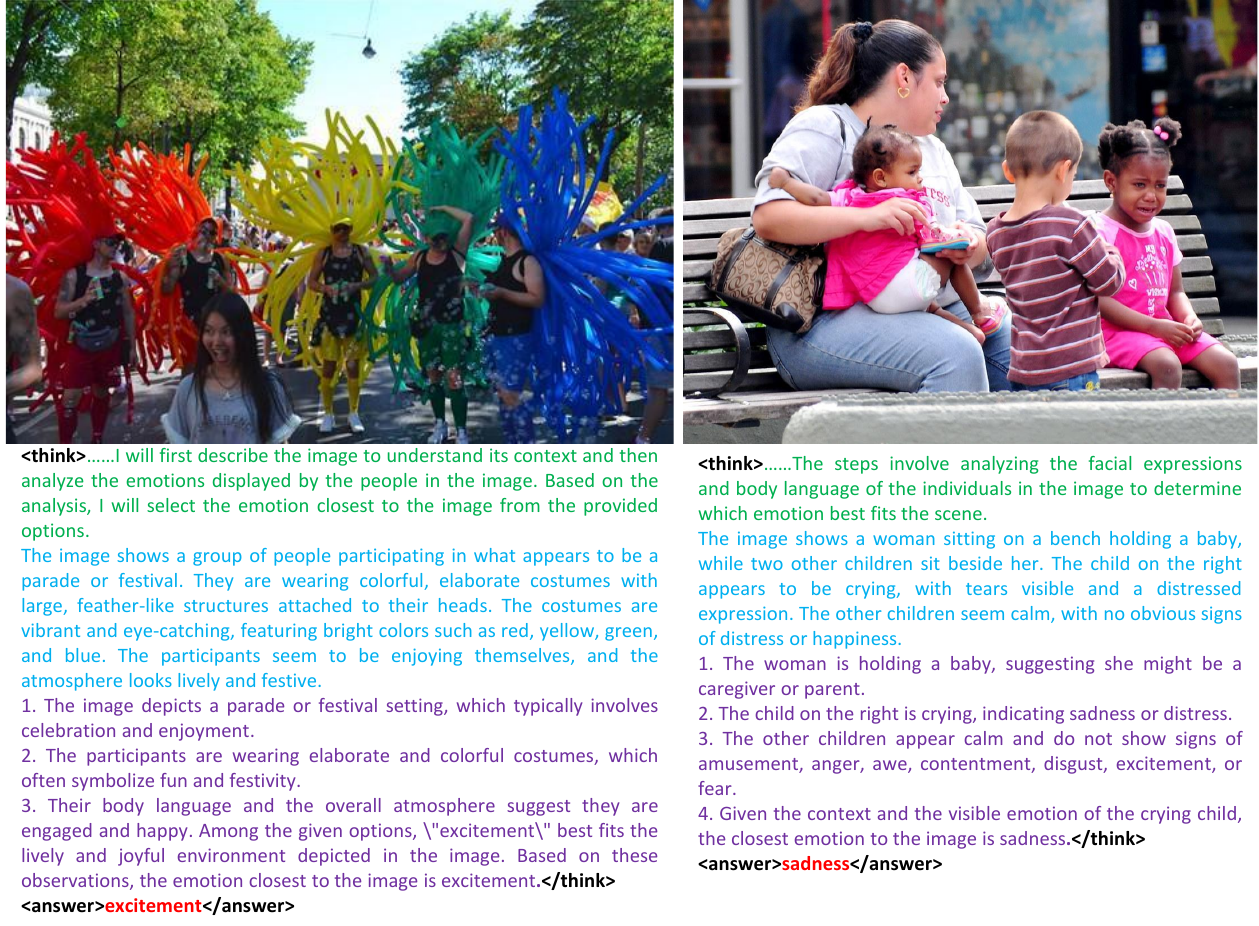}
    \captionof{figure}{COT Data produced by our Self-Supervised COT Data Construction System. \textcolor{Green}{Green is SUMMARY part}, \textcolor{Cyan}{Blue is CAPTION part}, \textcolor{Purple}{Purple is RESASONING part}, \textcolor{Red}{Red is CONCLUSION part}.}
    \label{fig:COTData}
\end{center}%
\end{figure}

\subsection{Incremental Training Strategy}
We chose the InternVL2.5-1B~\cite{wang2024internvl2.5} as the baseline to validate the effectiveness of our Incremental Training Strategy. The entire strategy is divided into four stages, as illustrated on the right of the Figure.~\ref{fig:Overview}. Each stage corresponds to a specific objective: Domain Knowledge Injection, Format Alignment, Reasoning Ability Enhancement, and Capture Space Constraint. Among these, Stage 2 and Stage 4 are specially designed to address the capability characteristics of SVLMs.

\subsubsection{Stage 1: Domain knowledge injection}
The purpose of the first training stage is to inject domain knowledge into the small model. For LVLMs, it has been validated that fine-tuning with COT data which contains richer information can improve their performance in the corresponding domain. However, for SVLMs, their limited parameter capacity constrain their COT capabilities to some extent. The COT reasoning process can be divided into two stages: knowledge extraction and reasoning. We believe that SVLMs can achieve performance comparable to LVLMs in vertical domains by first injecting domain knowledge and then enhancing reasoning ability.

The goal of this stage is to enable the SVLMs to learn to extract knowledge from domain-specific COT data. After the knowledge injection in the first stage, the SVLM has initially acquired domain-specific reasoning ability, enabling it to first understand image content, then analyze, and finally answer. However, as shown in Table.~\ref{tab:acc_on_emoset_comparison},~\ref{tab:recall_on_emoset_comparison}, there remains a gap in objective metrics compared to models fine-tuned directly on the original data, indicating the need for further enhancement of reasoning ability.
\subsubsection{Stage 2: Format Alignment}

Reinforcement learning is one of the core approaches to enhancing model reasoning capabilities. In our work, we leverage GRPO~\cite{deepseek-math} to improve the reasoning ability, where the feedback primarily consists of two components: format reward and accuracy reward. Compared to previous reinforcement learning algorithms~\cite{schulman2017ppo}\cite{rafailov2023dpo}, GRPO offers advantages such as eliminating the need for a separate value network, thereby reducing computational overhead, and employing group-structured advantage estimation, which leads to more stable training.

In practical experiments, we observed that SVLM exhibit weaker learning ability for the "$<$think$><$/think$><$answer$><$/answer$>$" format. The format reward fluctuates persistently during training and fails to reach 100\% accuracy. Consequently, some correctly formatted responses inadvertently become the focus of reinforcement learning process, which biases the model’s attention and causes confusion for the parameter-limited SVLM. Corresponding experimental results can be found in the ablation study, specifically in Table.~\ref{tab:ablation_stage2_acc_on_emoset_cot_sft},~\ref{tab:ablation_stage2_recall_on_emoset_cot_sft}.

\subsubsection{Stage 3: Reasoning Ability Enhancement}

After Stage 2, the model acquires the ability to respond in a fixed format. The focus of Stage 3 training process is Reasoning Ability Enhancement. This stage is also based on GRPO, with the corresponding objective function shown in Equation~\ref{GRPO}. 
\begin{equation}
\label{GRPO}
\begin{aligned}
\mathcal{L}_{\text{GRPO}} = \frac{1}{G}\sum_{i=1}^G \left.( \min \left.( \frac{\pi_\theta(o_i|q)}{\pi_{\theta_{\text{old}}}(o_i|q)} \cdot \text{A}_i, \text{clip}\left.(\frac{\pi_\theta(o_i|q)}{\pi_{\theta_{\text{old}}}(o_i|q)}, 1-\epsilon_{\text{l}}, 1+\epsilon_{\text{h}} ) \right. \cdot \text{A}_i )\right. + \beta \cdot \text{KL}(\pi_\theta \parallel \pi_{\text{ref}}) )\right.,\epsilon_{\text{l}}=\epsilon_{\text{h}}=0.2,\\
\end{aligned}
\end{equation}

where $\epsilon_l$ and $\epsilon_h$ are the clipping hyper-parameters, $\beta$ is the coefficient controlling the Kullback–Leibler (KL) penalty, $G$ is the group number, respectively, and $\text{A}_i = \frac{r(q, o_g) - mean(r_1,r_2,...,r_G)}{std(r_1,r_2,...,r_G)}$ is the computed advantage using the group rewards $\{r_1, r_2, · · · , r_G\}$. For each question $q$, GRPO samples a group of outputs $\{o_1, o_2, · · · , o_G\}$ from the old policy $\pi_{\theta_{\text{old}}}$ and then optimizes the policy model $\pi_{\theta}$ by maximizing the objective function above.

During Stage 3, both the format reward and accuracy reward are activated. The accuracy reward incentivizes the correctness of the model’s responses. Both rewards are positively correlated with the correctness of the model’s output. As shown in Table.~\ref{tab:acc_on_emoset_comparison},~\ref{tab:recall_on_emoset_comparison}, Stage 3 achieves significant improvements in accuracy and recall compared to Stage 1. However, compared to the model directly SFT on the original data , accuracy improves, but recall decreases. To analyze this phenomenon, the higher accuracy but lower recall indicates that the overall prediction reliability is high, but there are more false negatives on difficult samples. Considering the relatively weaker capability of SVLMs, we believe it is necessary to constrain the reinforcement learning capture space, enabling the small model to search more rigorously for correct answers within a smaller capture space. This insight motivated the design of the fourth training stage.

\subsubsection{Stage 4: Capture Space Constrain}
We achieve the goal of constraining the reinforcement learning capture space by restricting ${\epsilon}_l$ and ${\epsilon}_h$ to a smaller range. Based on empirical experiments, we adjust ${\epsilon}_l$ and $\epsilon_h$ from 0.2 in the stage 3 to 0.1, and conduct ablation studies to compare with the approach of increasing ${\epsilon}_h$ as mentioned in ByteDance’s DAPO~\cite{yu2025dapo}. The experimental results in Table.~\ref{tab:acc_on_emoset_comparison},~\ref{tab:recall_on_emoset_comparison} demonstrate the effectiveness of constraining the reinforcement learning capture space for SVLMs. To further verify that the performance gains are indeed due to the reduction of ${\epsilon}_l$ and ${\epsilon}_h$, we conducted additional ablation experiments shown in Table.~\ref{tab:ablation_cliplow_acc_on_emoset_cot_sft},~\ref{tab:ablation_cliplow_recall_on_emoset_cot_sft}, which also provide positive validation.


\begin{table*}[htbp]
\centering\small
  \begin{tabular}{ccccc} \hline
    \toprule
    &ORI Train Data  &COT Train Data &ORI/COT Test Data &ORI/COT Eval Data\\
    \midrule
    &94481 &94403 &17716 &5905\\
    \bottomrule
  \end{tabular}
 \caption{Data set partitioning}
\label{tab:DATASET}
\end{table*}

\begin{table*}[htbp]
\centering\scriptsize
  \begin{tabular}{cccccccccccc} \hline
    \toprule
    \textbf{Method}  &dataset &strategy &$amuse$ &$content$ &$excite$ &$awe$ &$disgust$ &$anger$ &$fear$ &$sad$ &$avg$\\
    \midrule
     InternVL2.5-1B  &- &- &37.49 &32.32  &61.08  &66.42 &72.06 &82.27 &48.65 &72.57 &59.11\\
    InternVL2.5-2B  &- &- &45.96 &35.92 &70.75  &65.31  &82.45 &84.13 &54.43 &69.76 &63.59 \\
    InternVL2.5-4B  &- &- &34.79 &48.67  &49.55 &54.55 &70.23 &87.84 &50.24 &81.46 &59.67  \\
    \midrule
    InternVL2.5-1B-a(baseline)  &ORI &SFT &69.42 &64.75  &80.97  &80.14 &87.40 &88.11 &76.78 &81.92 &78.68\\
    InternVL2.5-4B-a  &ORI &SFT & 73.02 &67.61  &81.44  &80.11 &87.59 &84.33 &81.09 &84.05 &79.91\\
    InternVL2.5-8B-a   &ORI &SFT &71.97 &67.74 &84.82 &80.03 &87.81 &88.72 &79.48 &84.10 &80.58\\
    InternVL2.5-1B-a-G  &ORI &GRPO &69.12 &64.94  &81.24  &79.54 &87.41 &89.10 &77.51 &82.86 &78.97\\
    InternVL2.5-1B-a-D    &ORI &Dr.GRPO &66.13 &67.11 &76.96 &91.11 &84.79 &85.36 &87.36 &71.38 &78.78 \\
    InternVL2.5-1B-b  &ORI &GRPO &67.18 &65.38  &79.35  &89.92 &85.81 &85.49 &82.75 &82.11 &79.75\\
    \midrule
    InternVL2.5-1B-c  &COT &GRPO &66.62 &54.84  &72.47  &71.46 &85.08 &86.98 &74.65 &80.54 &74.08\\
    InternVL2.5-1B(stage1)  &COT &SFT &64.40 &52.84  &73.30  &70.64 &84.86 &86.99 &73.86 &80.00 &73.36\\
    InternVL2.5-1B(stage2)  &COT &GRPO &64.18 &52.67  &73.26  &69.94 &83.41 &86.08 &74.96 &81.80 &73.29\\
    InternVL2.5-1B(stage3)  &COT &GRPO &80.00 &54.43  &77.31  &83.92 &88.17 &91.25 &83.60 &82.76 &80.31\\
    ~\textbf{InternVL2.5-1B(stage4-v1)}  &COT &CLGRPO &83.67 &58.49  &76.13  &86.90 &88.17 &91.86 &84.32 &82.09 &81.45\\
    InternVL2.5-1B(stage4-v2)  &COT &CLGRPO &82.89 &58.83  &76.68  &85.06 &88.37 &91.09 &83.64 &83.08 &81.20\\
    
    \bottomrule
  \end{tabular}
 \caption{Accuracy comparison experiments between different scale base models and different fintuning schedule models. In the Method column, the suffix “-a” indicates SFT on the ORI data. “-a-G” denotes GRPO training based on the SFT results “-a”. “-a-D” denotes Dr.GRPO training based on the SFT results “-a”. “-b” indicates direct GRPO training on the ORI data. and “-c” indicates direct GRPO training on the COT data.}
\label{tab:acc_on_emoset_comparison}
\end{table*}

\begin{table*}[htbp]
\centering\scriptsize
  \begin{tabular}{cccccccccccc} \hline
    \toprule
    \textbf{Method}  &dataset &strategy &$amuse$ &$content$ &$excite$ &$awe$ &$disgust$ &$anger$ &$fear$ &$sad$ &$avg$\\
    \midrule
    InternVL2.5-1B  &- &- &11.95 &90.98  &63.54  &36.15 &41.00 &40.75 &43.67 &50.19 &47.28\\
    InternVL2.5-2B  &- &- &25.27 &62.51 &62.51  &31.90  &55.15 &34.13 &49.75 &75.28 &53.05 \\
    InternVL2.5-4B  &- &- &17.00 &53.14 &67.78 &65.04 &54.97 &22.83 &31.66 &51.92 &45.54  \\
    \midrule
    InternVL2.5-1B-a(baseline)  &ORI &SFT &68.34 &65.14  &83.01  &79.47 &83.08 &88.06 &77.71 &83.25 &78.51\\
    InternVL2.5-4B-a  &ORI &SFT & 69.37 &66.65  &84.97  &82.06 &87.93 &89.34 &74.32 &86.10 &80.09\\
    InternVL2.5-8B-a  &ORI &SFT &71.08 &67.11  &84.17  &83.50 &86.33 &89.34 &79.48 &84.01 &80.63\\
    InternVL2.5-1B-a-G  &ORI &GRPO &67.08 &66.75  &83.48  &79.47 &84.89 &87.06 &77.86 &83.31 &78.74\\
    InternVL2.5-1B-a-D    &ORI &Dr.GRPO &63.65 &64.41 &88.98 &66.24 &87.24 &87.76 &67.58 &91.11 &77.12\\
    InternVL2.5-1B-b  &ORI &GRPO &63.39 &69.81  &84.51  &63.19 &80.79 &83.84 &70.47 &80.87 &74.61\\
    \midrule
    InternVL2.5-1B-c  &COT &GRPO &48.77 &72.26  &82.71  &74.12 &82.06 &71.19 &69.81 &77.83 &72.34\\
    InternVL2.5-1B(stage1)  &COT &SFT &50.87 & 72.30  &79.79  &74.29 & 78.99 &69.67 &67.83 &75.45 &71.15\\
    InternVL2.5-1B(stage2)  &COT &GRPO &51.23 & 71.86  &79.16  &72.79 & 80.49 &70.61 &67.33 &76.48 &71.24\\
    InternVL2.5-1B(stage3)  &COT &GRPO &51.70 &85.71  &88.29  &71.81 &87.96 &78.16 &71.53 &85.09 &77.53\\
    ~\textbf{InternVL2.5-1B(stage4-v1)}  &COT &CLGRPO &52.53 &83.21  &91.34  &73.36 &88.38 &81.91 &74.67 &88.18 &79.20\\
    InternVL2.5-1B(stage4-v2)  &COT &CLGRPO &52.31 &83.17  &90.54  &74.29 &88.26 &81.38 &75.38 &88.08 &79.18\\
    \bottomrule
  \end{tabular}
 \caption{Recall comparison experiments between different scale base models and different fintuning schedule models. In the Method column, the suffix “-a” indicates SFT on the ORI data. “-a-G” denotes GRPO training based on the SFT results “-a”. “-a-D” denotes Dr.GRPO training based on the SFT results “-a”. “-b” indicates direct GRPO training on the ORI data. and “-c” indicates direct GRPO training on the COT data.}
\label{tab:recall_on_emoset_comparison}
\end{table*}

\begin{table*}[htbp]
\centering\small
  \begin{tabular}{ccccccccccccc} \hline
    \toprule
    \textbf{Method}  &Stage2 &$amuse$ &$content$ &$excite$ &$awe$ &$disgust$ &$anger$ &$fear$ &$sad$ &$avg$\\
    \midrule
    InternVL2.5-1B(stage1)  &- &64.40 &52.84  &73.30  &70.64 &84.86 &86.99 &73.86 &80.00 &73.36\\
    InternVL2.5-1B(stage3)  &\XSolidBrush &67.16 &54.15  &75.98  &72.62 &84.71 &86.79 &75.70 &82.88 &75.00\\
    \textbf{InternVL2.5-1B(stage3)(ours)}  &\checkmark &80.00 &54.43  &77.31  &83.92 &88.17 &91.25 &83.60 &82.76 &80.31\\
    \bottomrule
  \end{tabular}
 \caption{Accuracy ablation study of Stage2 Format Alignment.}
\label{tab:ablation_stage2_acc_on_emoset_cot_sft}
\end{table*}

\begin{table*}[htbp]
\centering\small
  \begin{tabular}{ccccccccccccc} \hline
    \toprule
    \textbf{Method}  &Stage2 &$amuse$ &$content$ &$excite$ &$awe$ &$disgust$ &$anger$ &$fear$ &$sad$ &$avg$\\
    \midrule
    InternVL2.5-1B(stage1)   &- &50.87 & 72.30  &79.79  &74.29 & 78.99 &69.67 &67.83 &75.45 &71.15\\
    InternVL2.5-1B(stage3)  &\XSolidBrush &52.20 &73.51  &80.79  &73.58 &83.02 &73.48 &71.63 &78.48 &73.34\\
    \textbf{InternVL2.5-1B(stage3)(ours)}  &\checkmark &51.70 &85.71  &88.29  &71.81 &87.96 &78.16 &71.53 &85.09 &77.53\\
    \bottomrule
  \end{tabular}
 \caption{Recall ablation study of Stage2 Format Alignment.}
\label{tab:ablation_stage2_recall_on_emoset_cot_sft}
\end{table*}

\begin{figure}[ht]
\begin{center}
    \captionsetup{type=figure}
    \includegraphics[width=1.0\linewidth]{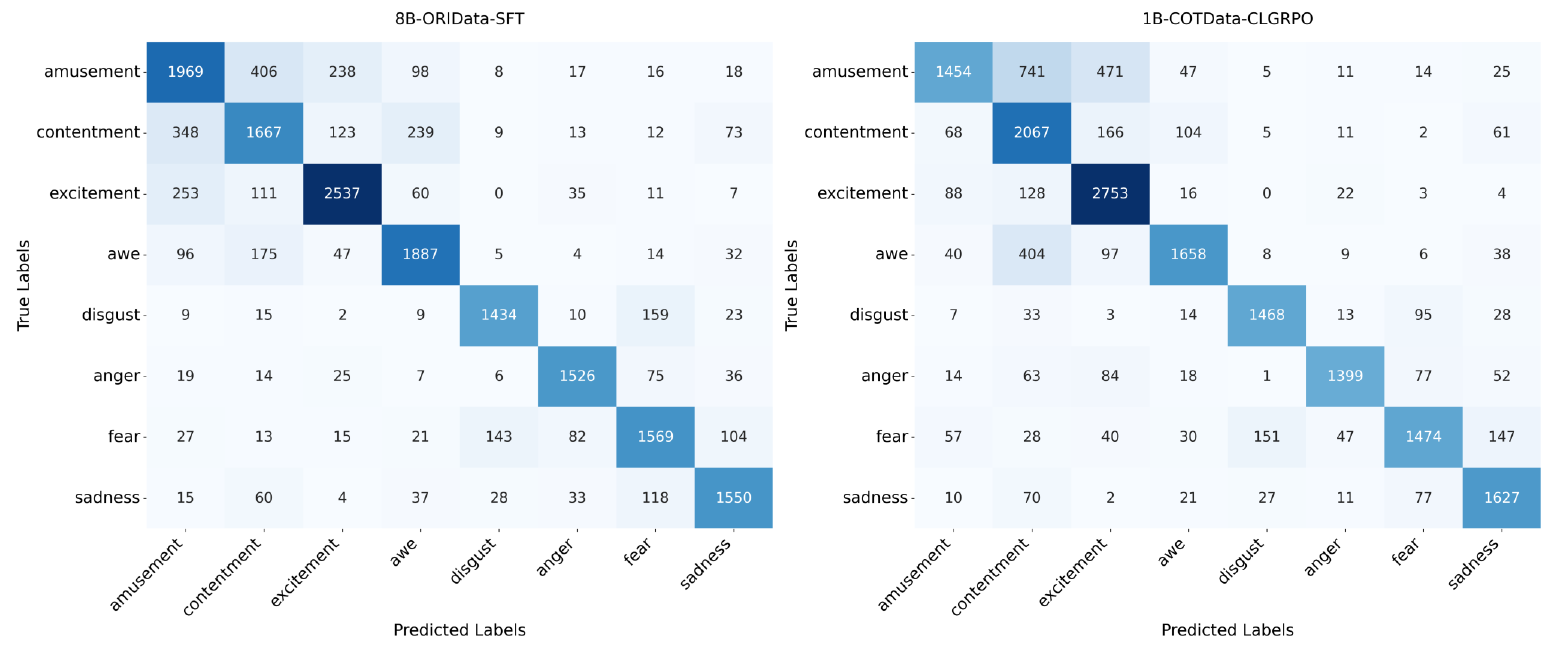}
    \captionof{figure}{Confusion matrix of InternVL-2.5-8B-Instruct SFT on ORI Data and our CLGRPO result train on COT Data.}
    \label{fig:Confusionmatrixs}
\end{center}%
\end{figure}

\begin{figure}[ht]
\begin{center}
    \captionsetup{type=figure}
    \includegraphics[width=0.85\linewidth]{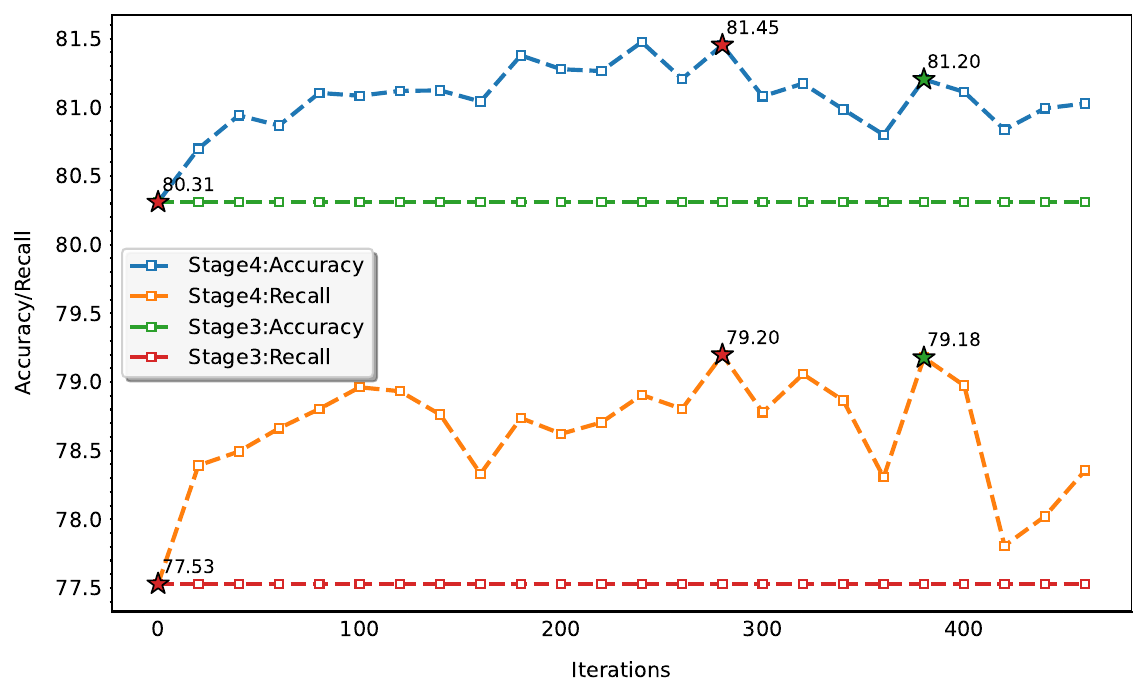}
    \captionof{figure}{Accuracy and recall metrics results during Stage4 training process compared with results of Stage3.}
    \label{fig:stage3vsstage4}
\end{center}%
\end{figure}

\section{Experiments}
This chapter first provides a detailed description of the training configurations for all comparative experiments, as well as the training configurations for the four stages of the proposed Incremental Training Strategy. Next, to validate the overall effectiveness of the Incremental Training Strategy, we conducted a series of comparative experiments. The comparative experiments, shown in Table.~\ref{tab:acc_on_emoset_comparison},~\ref{tab:recall_on_emoset_comparison}, including comparisons with InternVL-2.5-1B/2B/4B-Instruct base models; InternVL-2.5-1B/4B/8B fine-tuned on the original dataset; InternVL-2.5-1B trained with SFT followed by GRPO and Dr.GRPO~\cite{liu2025understanding} on the original dataset; InternVL-2.5-1B directly trained with GRPO on the original or COT dataset.

Furthermore, to verify the effectiveness of Stage 2 and Stage 4, which are specifically designed for the characteristics of SVLMs, we conducted two groups of ablation studies, as shown in Table.~\ref{tab:ablation_stage2_acc_on_emoset_cot_sft},~\ref{tab:ablation_stage2_recall_on_emoset_cot_sft}, and Table.~\ref{tab:ablation_cliplow_acc_on_emoset_cot_sft},~\ref{tab:ablation_cliplow_recall_on_emoset_cot_sft}, respectively. The first group ablates the execution of format alignment of Stage 2, while the second group ablates the values of $\epsilon_l$ and $\epsilon_h$. Through rigorous comparative and ablation experiments, we thoroughly validate that the proposed Incremental Training Strategy is indeed effective in enhancing the domain-specific COT capabilities of small Vision Language Models.

\subsection{Implementation details}

We selected InternVL2.5-1B-Instruct as the baseline. All experiments were conducted on 8$\times$A800 GPUs, and inferences were performed on 8$\times$L20 GPUs. All SFT experiments were trained for 3 epochs on either the ORI data or the COT data. For the 1B and 4B models, the training batch size was set to 8, while for the 8B model, the batch size was set to 2. Gradient accumulation steps were set to 8, weight decay to 0.05, warmup ratio to 0.03, and learning rate to 1e-5. During testing, the best-performing weights on the evaluation data were selected, and the testing data format was consistent with the training data format.
For all reinforcement learning experiments (GRPO, Dr.GRPO, DAPO), except for InternVL2.5-1B-b and InternVL2.5-1B-c in Table.~\ref{tab:acc_on_emoset_comparison},~\ref{tab:recall_on_emoset_comparison} which were trained for 3 epochs on the ORI data, the rest were trained for 1 epoch on either ORI or COT data. The batch size was set to 8, gradient accumulation steps to 8, group size to 8, $\beta$ to 0.04, and learning rate to 1e-6. The default values of $\epsilon_l$ and $\epsilon_h$ in GRPO were 0.2, while in ClipLow GRPO (CLGRPO), both were set to 0.1.

In the four-stage Incremental Training Strategy, each stage was trained for 3 epochs, 10 iterations, 1 epoch, and 500 iterations respectively. In Stages 2 and 3, $\epsilon_l$ and $\epsilon_h$ were set to 0.2, while in Stage 4, both were set to the empirical value of 0.1. The experimental data is EmoSet-118K~\cite{EmoSet}. During the COT data construction process, 78 samples were filtered out by the Self-Supervised COT Data Construction System. The data were split into training, testing, and evaluation sets in a 16:3:1 ratio, with specific data volumes shown in Table~\ref{tab:DATASET}. The evaluation and test data volumes for both ORI and COT formats were consistent, while the ORI training data contained 78 more samples than the COT training data.

\subsection{Comparison Experiments}

To validate the effectiveness of the Incremental Training Strategy, we conducted a series of comparative experiments as shown in Table.~\ref{tab:acc_on_emoset_comparison},~\ref{tab:recall_on_emoset_comparison}. The experiments are divided into two groups based on whether training is involved: tuning-free and tuning-based control groups. In the Method column of Table.~\ref{tab:acc_on_emoset_comparison},~\ref{tab:recall_on_emoset_comparison}, the first row corresponds to tuning-free experiments, the second row to tuning-based experiments, and the third row to the four stages of the proposed Strategy. For Stage 4, both the best and second-best results are listed.

In the tuning-free control group, we directly tested the base models on the test data split described in Table.~\ref{tab:DATASET} without any domain-specific fine-tuning. These models performed relatively poorly. For the tuning-based control group, training and testing were conducted on the ORI training and test data, with the best-performing weights selected based on evaluation data performance.  Since training was performed on ORI data without format constraints, GRPO training in the tuning-based experiments used only the accuracy reward for supervision. In our Incremental Training Strategy experimental group, Stage 1 involved SFT training on the COT data, Stage 2 focused solely on format alignment training on the COT data, and Stage 3 performed full GRPO training on the COT data, resulting in significant metric improvements. From this, we draw the preliminary conclusion that by Stage 3, the model has acquired initial COT reasoning ability, achieving performance comparable to that of an 8B LVLM trained for 3 epochs on ORI data.

However, not satisfied with this, and based on the analysis of SVLMs characteristics in the previous chapter, we further designed the Stage 4 experiment. The two best-performing weights on the test data are listed in Table.~\ref{tab:acc_on_emoset_comparison},~\ref{tab:recall_on_emoset_comparison}. It can be observed that the final results show significant improvement over Stage 3. Compared to the 8B model trained for 3 epochs on ORI data, the accuracy metric is superior, while recall shows some decline. Figure.~\ref{fig:Confusionmatrixs} illustrates the confusion matrix of InternVL-8B-Instruct trained on the original data and best results of Stage4. Based on the experimental results in Table.~\ref{tab:acc_on_emoset_comparison},~\ref{tab:recall_on_emoset_comparison}, we summarize the observed phenomena on the EmoSet benchmark for SVLM as follows:

\begin{itemize}
\item The GRPO training results show higher accuracy compared to SFT on the original data, but recall decreases.\textbf{(InternVL2.5-1B-a(baseline) vs InternVL2.5-1B-b)}
\item Performing SFT as a cold start before GRPO training further improves the metrics. Compared to directly training with GRPO, this approach achieves a more balanced accuracy and recall, with a significant improvement in recall.\textbf{(InternVL2.5-1B-a-G vs InternVL2.5-1B-b)}
\item Dr.GRPO, originally validated for mathematical reasoning ability, performs worse than GRPO when transferred to the visual abstract semantic understanding task according to experimental results.\textbf{(InternVL2.5-1B-a-G vs InternVL2.5-1B-a-D)}
\item Without knowledge injection, directly training with GRPO on the COT data results in a significant performance drop compared to directly training with GRPO on the original data.\textbf{(InternVL2.5-1B-b vs InternVL2.5-1B-c)}
\end{itemize}

\begin{table*}[htb]
\centering\scriptsize
  \begin{tabular}{ccccccccccccc} \hline
    \toprule
    \textbf{Method}  &strategy &$\epsilon_l,\epsilon_h$ &$amuse$ &$content$ &$excite$ &$awe$ &$disgust$ &$anger$ &$fear$ &$sad$ &$avg$\\
    \midrule
    InternVL2.5-1B(stage1)  &SFT &- &64.40 &52.84  &73.30  &70.64 &84.86 &86.99 &73.86 &80.00 &73.36\\
    InternVL2.5-1B(stage3)  &GRPO &0.2,0.2 &80.00 &54.43  &77.31  &83.92 &88.17 &91.25 &83.60 &82.76 &80.31\\
    InternVL2.5-1B(stage4)  &GRPO &0.2,0.28 &76.48 &56.44  &77.02  &85.17 &87.84 &90.92 &82.04 &82.19 &79.76\\
    \textbf{InternVL2.5-1B(stage4)(ours)}  &CLGRPO &0.1,0.1 &83.67 &58.49  &76.13  &86.90 &88.17 &91.86 &84.32 &82.09 &81.45\\
    
    \bottomrule
  \end{tabular}
 \caption{Accuracy ablation study of Stage4 CLGRPO.}
\label{tab:ablation_cliplow_acc_on_emoset_cot_sft}
\end{table*}

\begin{table*}[htb]
\centering\scriptsize
  \begin{tabular}{ccccccccccccc} \hline
    \toprule
    \textbf{Method}  &strategy &$\epsilon_l,\epsilon_h$ &$amuse$ &$content$ &$excite$ &$awe$ &$disgust$ &$anger$ &$fear$ &$sad$ &$avg$\\
    \midrule
    InternVL2.5-1B(stage1)  &SFT &- &50.87 & 72.30  &79.79  &74.29 & 78.99 &69.67 &67.83 &75.45 &71.15\\
    InternVL2.5-1B(stage3)  &GRPO &0.2,0.2 &51.70 &85.71  &88.29  &71.81 &87.96 &78.16 &71.53 &85.09 &77.53\\
    InternVL2.5-1B(stage4) &GRPO &0.2,0.28 &53.75 &84.46  &87.06  &71.90 &86.51 &77.93 &72.44 &84.28 &77.29\\
    \textbf{InternVL2.5-1B(stage4)(ours)}  &CLGRPO &0.1,0.1 &52.53 &83.21  &91.34  &73.36 &88.38 &81.91 &74.67 &88.18 &79.20\\
    \bottomrule
  \end{tabular}
 \caption{Recall ablation study of Stage4 CLGRPO.}
\label{tab:ablation_cliplow_recall_on_emoset_cot_sft}
\end{table*}

\subsection{Ablation Experiments}

To verify the necessity of the Stage 2 and Stage 4 training phases designed specifically for small Vision Language Models, we conducted two groups of ablation studies. The experimental results are shown in Table.~\ref{tab:ablation_stage2_acc_on_emoset_cot_sft},~\ref{tab:ablation_stage2_recall_on_emoset_cot_sft} and Table.~\ref{tab:ablation_cliplow_acc_on_emoset_cot_sft},~\ref{tab:ablation_cliplow_recall_on_emoset_cot_sft}, respectively.

First, the purpose of Stage 2 is to enhance the format alignment capability of SVLMs and avoid confusion during training process. In Table.~\ref{tab:ablation_stage2_acc_on_emoset_cot_sft},~\ref{tab:ablation_stage2_recall_on_emoset_cot_sft}, the second-to-last row shows the results of directly proceeding to Stage 3 without executing Stage 2, where GRPO was trained using both format reward and accuracy reward. Although this shows improvement compared to the Stage 1 baseline, there remains a performance gap compared to models fine-tuned on the ORI data as shown in Table.~\ref{tab:acc_on_emoset_comparison},~\ref{tab:recall_on_emoset_comparison}. The last row presents the results of performing Stage 2 with only format reward constrained training for 10 iterations, followed by Stage 3 training. It can be observed that the accuracy metric significantly improves compared to baseline trained on ORI data. These results validate that, for SVLMs, it is necessary to perform format alignment before enhancing reasoning ability. Due to training resource limitations, we did not verify whether larger models will benifit from format alignment pre-training.


To verify the effectiveness of ClipLow operation in stage 4, we performed ablation experiments as shown in Table~\ref{tab:ablation_cliplow_acc_on_emoset_cot_sft},~\ref{tab:ablation_cliplow_recall_on_emoset_cot_sft}. And Figure.~\ref{fig:stage3vsstage4} illustrates the change in metrics evaluated in the test data during the training process of Stage 4. In Stage 3, $\epsilon_l$ and $\epsilon_h$ were set to 0.2. We performed a set of experiments increasing the thresholds $\epsilon_l$ and $\epsilon_h$ , with values referenced from DAPO~\cite{yu2025dapo} where $\epsilon_l$=0.2 and $\epsilon_h$=0.28. These experiments were conducted by training for 500 iterations on the best weights obtained from Stage 3. From the experimental results, it can be observed that, as shown in the second-to-last row, setting $\epsilon_h$ to 0.28 leads to a decline in metrics compared to the Stage 3. The last row presents the results of our proposed ClipLow operation tailored for SVLMs, which shows significant improvement over the Stage 3. This experimental phenomenon further supports our hypothesis that constraining the capture space of SVLMs to reduce the search difficulty can further enhance the Chain-of-Thought (COT) reasoning ability of small models in vertical domain tasks.

\section{Limitations}
Due to computational resource limitations, our experiments were conducted only on the abstract semantic understanding task using the EmoSet-118K benchmark~\cite{EmoSet}, and we have not yet validated the effectiveness of the Incremental Training Strategy on other SVLMs. Additionally, there is a lack of in-depth analysis of some phenomena observed in the experiments. For example, in Table.~\ref{tab:acc_on_emoset_comparison},~\ref{tab:recall_on_emoset_comparison}, for the emotion subclass "amuse", our proposed method achieves a significant improvement in accuracy compared to the model trained on ORI data, but recall decreases noticeably. We speculate that this emotion category is more prone to confusion. During the Stage 1 knowledge injection phase, the SVLM may not clearly delineate boundaries between closely related concepts. This hypothesis is supported by the experimental results showing that SFT LVLMs have a clear advantage in recall. However, we currently lack concrete experimental validation for this point.

\section{Analysis and Conclusions}
This paper posits that SVLMS can also achieve reasoning capabilities comparable to LVLMs through stepwise domain-specific knowledge injection and reasoning ability enhancement. We propose an Incremental Training Strategy specifically designed to improve the reasoning ability of SVLMs, and validate it experimentally on the abstract semantic sentiment recognition dataset EmoSet-118K. Additionally, we developed a Self-Supervised COT Data Construction System to automatically convert the EmoSet dataset into a Chain-of-Thought formatted data type. Comparative experimental results demonstrate that our proposed Incremental Training Strategy significantly enhances the reasoning ability of SVLM, InternVL2.5-1B-Instruct, achieving performance metrics comparable to those of 8B SFT models. Ablation studies on Stage 2 indicate that a small amount of format alignment training for SVLM can substantially improve reinforcement learning-based reasoning capabilities. Ablation studies on Stage 4 show that constraining the capture space during reinforcement learning to reduce search difficulty enables SVLMs to more accurately locate better solutions within a smaller search space. We hope our work serves as a pioneering effort to attract more researchers to the study of SVLMs, thereby empowering industrial applications such as autonomous driving on edge devices.





\bibliographystyle{ijcai25}


\begin{thebibliography}{10}
\providecommand{\url}[1]{#1}
\csname url@samestyle\endcsname
\providecommand{\newblock}{\relax}
\providecommand{\bibinfo}[2]{#2}
\providecommand{\BIBentrySTDinterwordspacing}{\spaceskip=0pt\relax}
\providecommand{\BIBentryALTinterwordstretchfactor}{4}
\providecommand{\BIBentryALTinterwordspacing}{\spaceskip=\fontdimen2\font plus
\BIBentryALTinterwordstretchfactor\fontdimen3\font minus \fontdimen4\font\relax}
\providecommand{\BIBforeignlanguage}[2]{{%
\expandafter\ifx\csname l@#1\endcsname\relax
\typeout{** WARNING: IEEEtran.bst: No hyphenation pattern has been}%
\typeout{** loaded for the language `#1'. Using the pattern for}%
\typeout{** the default language instead.}%
\else
\language=\csname l@#1\endcsname
\fi
#2}}
\providecommand{\BIBdecl}{\relax}
\BIBdecl

\bibitem{Qwen2.5-VL}
S.~Bai, K.~Chen, X.~Liu, J.~Wang, W.~Ge, S.~Song, K.~Dang, P.~Wang, S.~Wang, J.~Tang, H.~Zhong, Y.~Zhu, M.~Yang, Z.~Li, J.~Wan, P.~Wang, W.~Ding, Z.~Fu, Y.~Xu, J.~Ye, X.~Zhang, T.~Xie, Z.~Cheng, H.~Zhang, Z.~Yang, H.~Xu, and J.~Lin, ``Qwen2.5-vl technical report,'' \emph{arXiv preprint arXiv:2502.13923}, 2025.

\bibitem{Qwen2-VL}
P.~Wang, S.~Bai, S.~Tan, S.~Wang, Z.~Fan, J.~Bai, K.~Chen, X.~Liu, J.~Wang, W.~Ge, Y.~Fan, K.~Dang, M.~Du, X.~Ren, R.~Men, D.~Liu, C.~Zhou, J.~Zhou, and J.~Lin, ``Qwen2-vl: Enhancing vision-language model's perception of the world at any resolution,'' \emph{arXiv preprint arXiv:2409.12191}, 2024.

\bibitem{Qwen-VL}
J.~Bai, S.~Bai, S.~Yang, S.~Wang, S.~Tan, P.~Wang, J.~Lin, C.~Zhou, and J.~Zhou, ``Qwen-vl: A versatile vision-language model for understanding, localization, text reading, and beyond,'' \emph{arXiv preprint arXiv:2308.12966}, 2023.

\bibitem{zhu2025internvl3exploringadvancedtraining}
\BIBentryALTinterwordspacing
J.~Zhu, W.~Wang, Z.~Chen, Z.~Liu, S.~Ye, L.~Gu, H.~Tian, Y.~Duan, W.~Su, J.~Shao, Z.~Gao, E.~Cui, X.~Wang, Y.~Cao, Y.~Liu, X.~Wei, H.~Zhang, H.~Wang, W.~Xu, H.~Li, J.~Wang, N.~Deng, S.~Li, Y.~He, T.~Jiang, J.~Luo, Y.~Wang, C.~He, B.~Shi, X.~Zhang, W.~Shao, J.~He, Y.~Xiong, W.~Qu, P.~Sun, P.~Jiao, H.~Lv, L.~Wu, K.~Zhang, H.~Deng, J.~Ge, K.~Chen, L.~Wang, M.~Dou, L.~Lu, X.~Zhu, T.~Lu, D.~Lin, Y.~Qiao, J.~Dai, and W.~Wang, ``Internvl3: Exploring advanced training and test-time recipes for open-source multimodal models,'' 2025. [Online]. Available: \url{https://arxiv.org/abs/2504.10479}
\BIBentrySTDinterwordspacing

\bibitem{gao2024mini}
Z.~Gao, Z.~Chen, E.~Cui, Y.~Ren, W.~Wang, J.~Zhu, H.~Tian, S.~Ye, J.~He, X.~Zhu \emph{et~al.}, ``Mini-internvl: a flexible-transfer pocket multi-modal model with 5\% parameters and 90\% performance,'' \emph{Visual Intelligence}, vol.~2, no.~1, pp. 1--17, 2024.

\bibitem{chen2024internvl}
Z.~Chen, J.~Wu, W.~Wang, W.~Su, G.~Chen, S.~Xing, M.~Zhong, Q.~Zhang, X.~Zhu, L.~Lu \emph{et~al.}, ``Internvl: Scaling up vision foundation models and aligning for generic visual-linguistic tasks,'' in \emph{Proceedings of the IEEE/CVF Conference on Computer Vision and Pattern Recognition}, 2024, pp. 24\,185--24\,198.

\bibitem{deepseekai2024deepseekv3technicalreport}
\BIBentryALTinterwordspacing
DeepSeek-AI, ``Deepseek-v3 technical report,'' 2024. [Online]. Available: \url{https://arxiv.org/abs/2412.19437}
\BIBentrySTDinterwordspacing

\bibitem{deepseekai2025deepseekr1incentivizingreasoningcapability}
\BIBentryALTinterwordspacing
------, ``Deepseek-r1: Incentivizing reasoning capability in llms via reinforcement learning,'' 2025. [Online]. Available: \url{https://arxiv.org/abs/2501.12948}
\BIBentrySTDinterwordspacing

\bibitem{wu2024deepseekvl2mixtureofexpertsvisionlanguagemodels}
\BIBentryALTinterwordspacing
Z.~Wu, X.~Chen, Z.~Pan, X.~Liu, W.~Liu, D.~Dai, H.~Gao, Y.~Ma, C.~Wu, B.~Wang, Z.~Xie, Y.~Wu, K.~Hu, J.~Wang, Y.~Sun, Y.~Li, Y.~Piao, K.~Guan, A.~Liu, X.~Xie, Y.~You, K.~Dong, X.~Yu, H.~Zhang, L.~Zhao, Y.~Wang, and C.~Ruan, ``Deepseek-vl2: Mixture-of-experts vision-language models for advanced multimodal understanding,'' 2024. [Online]. Available: \url{https://arxiv.org/abs/2412.10302}
\BIBentrySTDinterwordspacing

\bibitem{lu2024deepseekvl}
H.~Lu, W.~Liu, B.~Zhang, B.~Wang, K.~Dong, B.~Liu, J.~Sun, T.~Ren, Z.~Li, H.~Yang, Y.~Sun, C.~Deng, H.~Xu, Z.~Xie, and C.~Ruan, ``Deepseek-vl: Towards real-world vision-language understanding,'' 2024.

\bibitem{fastvlm2025}
P.~K.~A. Vasu, F.~Faghri, C.-L. Li, C.~Koc, N.~True, A.~Antony, G.~Santhanam, J.~Gabriel \emph{et~al.}, ``Fastvlm: Efficient vision encoding for vision language models,'' in \emph{Proceedings of the IEEE/CVF Conference on Computer Vision and Pattern Recognition (CVPR)}, June 2025.

\bibitem{EmoSet}
J.~Yang, Q.~Huang, T.~Ding, D.~Lischinski, D.~Cohen-Or, and H.~Huang, ``Emoset: A large-scale visual emotion dataset with rich attributes,'' in \emph{ICCV}, 2023.

\bibitem{chen2024internvl1.5}
Z.~Chen, W.~Wang, H.~Tian, S.~Ye, Z.~Gao, E.~Cui, W.~Tong, K.~Hu, J.~Luo, Z.~Ma \emph{et~al.}, ``How far are we to gpt-4v? closing the gap to commercial multimodal models with open-source suites,'' \emph{Science China Information Sciences}, vol.~67, no.~12, p. 220101, 2024.

\bibitem{deepseek-math}
\BIBentryALTinterwordspacing
Z.~Shao, P.~Wang, Q.~Zhu, R.~Xu, J.~Song, M.~Zhang, Y.~Li, Y.~Wu, and D.~Guo, ``Deepseekmath: Pushing the limits of mathematical reasoning in open language models,'' 2024. [Online]. Available: \url{https://arxiv.org/abs/2402.03300}
\BIBentrySTDinterwordspacing

\bibitem{yu2025dapo}
Q.~Yu, Z.~Zhang, R.~Zhu, Y.~Yuan, X.~Zuo, Y.~Yue, T.~Fan, G.~Liu, L.~Liu, X.~Liu \emph{et~al.}, ``Dapo: An open-source llm reinforcement learning system at scale,'' \emph{arXiv preprint arXiv:2503.14476}, 2025.

\bibitem{chu2023mobilevlm}
X.~Chu, L.~Qiao, X.~Lin, S.~Xu, Y.~Yang, Y.~Hu, F.~Wei, X.~Zhang, B.~Zhang, X.~Wei \emph{et~al.}, ``Mobilevlm: A fast, reproducible and strong vision language assistant for mobile devices,'' \emph{arXiv preprint arXiv:2312.16886}, 2023.

\bibitem{wang2024internvl2.5}
W.~Wang, Z.~Chen, W.~Wang, Y.~Cao, Y.~Liu, Z.~Gao, J.~Zhu, X.~Zhu, L.~Lu, Y.~Qiao, and J.~Dai, ``Enhancing the reasoning ability of multimodal large language models via mixed preference optimization,'' \emph{arXiv preprint arXiv:2411.10442}, 2024.

\bibitem{zhang2025flashvl}
B.~Zhang, S.~Li, R.~Tian, Y.~Yang, J.~Tang, J.~Zhou, and L.~Ma, ``Flash-vl 2b: Optimizing vision-language model performance for ultra-low latency and high throughput,'' \emph{arXiv preprint arXiv:2505.09498}, 2025.

\bibitem{schulman2017ppo}
J.~Schulman, F.~Wolski, P.~Dhariwal, A.~Radford, and O.~Klimov, ``Proximal policy optimization algorithms,'' \emph{arXiv preprint arXiv:1707.06347}, 2017.

\bibitem{rafailov2023dpo}
\BIBentryALTinterwordspacing
R.~Rafailov, A.~Sharma, E.~Mitchell, C.~D. Manning, S.~Ermon, and C.~Finn, ``Direct preference optimization: Your language model is secretly a reward model,'' in \emph{Thirty-seventh Conference on Neural Information Processing Systems}, 2023. [Online]. Available: \url{https://arxiv.org/abs/2305.18290}
\BIBentrySTDinterwordspacing

\bibitem{2023bluelm}
B.~Team, ``Bluelm: An open multilingual 7b language model,'' \url{https://github.com/vivo-ai-lab/BlueLM}, 2023.

\bibitem{yang2025qwen3}
A.~Yang, A.~Li, B.~Yang, B.~Zhang, B.~Hui, B.~Zheng, B.~Yu, C.~Gao, C.~Huang, C.~Lv \emph{et~al.}, ``Qwen3 technical report,'' \emph{arXiv preprint arXiv:2505.09388}, 2025.

\bibitem{chen2024expanding}
Z.~Chen, W.~Wang, Y.~Cao, Y.~Liu, Z.~Gao, E.~Cui, J.~Zhu, S.~Ye, H.~Tian, Z.~Liu \emph{et~al.}, ``Expanding performance boundaries of open-source multimodal models with model, data, and test-time scaling,'' \emph{arXiv preprint arXiv:2412.05271}, 2024.

\bibitem{shao2024visual}
H.~Shao, S.~Qian, H.~Xiao, G.~Song, Z.~Zong, L.~Wang, Y.~Liu, and H.~Li, ``Visual cot: Advancing multi-modal language models with a comprehensive dataset and benchmark for chain-of-thought reasoning,'' \emph{Advances in Neural Information Processing Systems}, vol.~37, pp. 8612--8642, 2024.

\bibitem{zhao2025cot}
Q.~Zhao, Y.~Lu, M.~J. Kim, Z.~Fu, Z.~Zhang, Y.~Wu, Z.~Li, Q.~Ma, S.~Han, C.~Finn \emph{et~al.}, ``Cot-vla: Visual chain-of-thought reasoning for vision-language-action models,'' in \emph{Proceedings of the Computer Vision and Pattern Recognition Conference}, 2025, pp. 1702--1713.

\bibitem{schulman2017proximal}
J.~Schulman, F.~Wolski, P.~Dhariwal, A.~Radford, and O.~Klimov, ``Proximal policy optimization algorithms,'' \emph{arXiv preprint arXiv:1707.06347}, 2017.

\bibitem{rafailov2023direct}
R.~Rafailov, A.~Sharma, E.~Mitchell, C.~D. Manning, S.~Ermon, and C.~Finn, ``Direct preference optimization: Your language model is secretly a reward model,'' \emph{Advances in Neural Information Processing Systems}, vol.~36, pp. 53\,728--53\,741, 2023.

\bibitem{su2025reveal}
X.~Su, Y.~Wang, J.~Zhu, M.~Yi, F.~Xu, Z.~Ma, and Y.~Liu, ``Reveal the mystery of dpo: The connection between dpo and rl algorithms,'' \emph{arXiv preprint arXiv:2502.03095}, 2025.

\bibitem{liu2025understanding}
Z.~Liu, C.~Chen, W.~Li, P.~Qi, T.~Pang, C.~Du, W.~S. Lee, and M.~Lin, ``Understanding r1-zero-like training: A critical perspective,'' \emph{arXiv preprint arXiv:2503.20783}, 2025.

\bibitem{xu2023u}
J.~Xu, L.~Xu, Y.~Yang, X.~Li, F.~Wang, Y.~Xie, Y.-J. Huang, and Y.~Li, ``u-llava: Unifying multi-modal tasks via large language model,'' \emph{arXiv preprint arXiv:2311.05348}, 2023.

\bibitem{xu2025reproducibility}
J.~Xu, X.~Wang, L.~Xu, Y.~Yang, X.~Li, F.~Wang, Y.~Xie, Y.-J. Huang, Y.~Li, and Y.~Hu, ``Reproducibility companion paper: u-llava: Unifying multi-modal tasks via large language model,'' in \emph{Proceedings of the 2025 International Conference on Multimedia Retrieval}, 2025, pp. 1964--1967.

\bibitem{wang2025overcoming}
A.~Wang, Z.~Zhang, D.~Wang, F.~Wang, H.~Hu, J.~Guo, Y.~Zhou, C.~Pang, and S.~Wen, ``Overcoming heterogeneous data in federated medical vision-language pre-training: A triple-embedding model selector approach,'' in \emph{Proceedings of the AAAI Conference on Artificial Intelligence}, vol.~39, no.~7, 2025, pp. 7500--7508.

\end{thebibliography}

\end{document}